%% file: latex/acl_latex.tex
\documentclass[11pt]{article}

\usepackage[final]{acl}

\usepackage{times}
\usepackage{latexsym}
\usepackage{amsmath}

\usepackage[T1]{fontenc}
\usepackage[utf8]{inputenc}

\usepackage{microtype}

\usepackage{inconsolata}

\usepackage{graphicx}

\usepackage{booktabs} % 也就是三线表
\usepackage{colortbl} % 用于表格着色
\usepackage{xcolor}   % 颜色支持
\usepackage{amssymb}
\usepackage{multirow}
\usepackage{bm}
\usepackage{algorithm}
\usepackage{algpseudocode}
\usepackage{arydshln}

\title{VocalNet-MDM: Accelerating Streaming Speech LLM via Self-Distilled Masked Diffusion Modeling}

\author{%
Ziyang Cheng$^{1,2\ast}$ \quad
Yuhao Wang$^{1\ast}$ \quad
Heyang Liu$^{1,2}$ \quad
Ronghua Wu$^{2}$ \quad
Qunshan Gu$^{2}$ \\
\textbf{Yanfeng Wang}$^{1}$ \quad
\textbf{Yu Wang}$^{1\dagger}$\\
$^{1}$Shanghai Jiao Tong University \quad
$^{2}$ Ant Group\\
\texttt{\{colane, liuheyang, wangyanfeng622, yuwangsjtu, muye12\}@sjtu.edu.cn}\\
\texttt{\{r.wu, guqunshan.gqs\}@antgroup.com}
}

\begin{document}
\maketitle
\begin{abstract}
Recent Speech Large Language Models~(LLMs) have achieved impressive capabilities in end-to-end speech interaction. However, the prevailing autoregressive paradigm imposes strict serial constraints, limiting generation efficiency and introducing exposure bias. In this paper, we investigate Masked Diffusion Modeling~(MDM) as a non-autoregressive paradigm for speech LLMs and introduce VocalNet-MDM. To adapt MDM for streaming speech interaction, we address two critical challenges: 
training-inference mismatch and iterative overhead.
We propose Hierarchical Block-wise Masking to align training objectives with the progressive masked states encountered during block diffusion decoding, and Iterative Self-Distillation to compress multi-step refinement into fewer steps for low-latency inference. Trained on a limited scale of only 6K hours of speech data, VocalNet-MDM achieves a 3.7$\times$--10$\times$ decoding speedup and reduces first-chunk latency by 34\% compared to AR baselines. It maintains competitive recognition accuracy while achieving state-of-the-art text quality and speech naturalness, demonstrating that MDM is a promising and scalable alternative for low-latency, efficient speech LLMs.
\end{abstract}

\input{sections/1_Introduction}
\input{sections/2_Preliminary}
\input{sections/3_Methodology}
\input{sections/4_Experiments}
\input{sections/5_Conclusion}

\section*{Limitations}
Several aspects present opportunities for future improvement. Our training data scale remains modest compared to recent large-scale models, suggesting potential performance gains with expanded corpora. While distillation effectively reduces inference steps, quality-efficiency trade-offs become more pronounced at extremely low step counts. Speech-text alignment performance, though competitive with most baselines, leaves room for refinement through advanced optimization techniques. We alse leave exploring how bidirectional context modeling affects speech quality for future work. According to results in Section~\ref{subsec:detailed_efficiency_analysis}, vocoder latency constitutes the dominant portion of end-to-end delay due to flow matching constraints, indicating that further system-level acceleration requires joint optimization across all components. Future work may explore larger-scale pretraining, enhanced distillation strategies, and integrated vocoder acceleration.

\section*{Ethical Considerations}
All pre-trained models used in this work, including Qwen3-8B, Whisper-large-v3, and CosyVoice2, were obtained from publicly available sources and used in strict compliance with their respective licenses. The datasets employed in our experiments, including VoiceAssistant-400K and UltraChat, are publicly available and were used according to their specified terms. We did not collect any original data for this study. The speech data utilized in our experiments were either sourced from publicly available datasets or synthesized using the open-source CosyVoice2 text-to-speech model based on these datasets, minimizing potential risks related to privacy, consent, and data misuse. By relying on established, ethically sourced data and models without accessing private or sensitive information, we ensure that our research adheres to responsible AI practices throughout the development and evaluation process.

\bibliography{custom}

\appendix

\input{sections/6_Appendix}

\end{document}

%% file: sections/1_Introduction.tex
\section{Introduction}

Recent Speech LLMs have advanced end-to-end speech interaction through diverse multimodal architectures~\citep{zeng2024glm,coreteam2025mimoaudio,li2025baichuan,wu2025step,Qwen3-Omni,wang2025vocalnet1,xu2025qwen2}.
However, reliance on the Autoregressive (AR) paradigm limits real-time interaction. Latency grows linearly with sequence length due to serial dependence, and exposure bias degrades quality~\citep{lin2025accelerating,fan2025makesgoodspeechtokenizer,chen2025f5}.
Despite various explorations to mitigate these issues~\citep{long2025vita,wang2025vocalnet1,wang2025vocalnet,bhati2025towards}, 
they remain within the AR framework, preventing substantial throughput improvements.

% We further ask: does speech generation truly require strict adherence to the serial constraint of AR decoding?
% We further ask: does speech generation truly require strict adherence to the strictly serial, AR formulation?
% This naturally motivates non-autoregressive (NAR) generation, which predicts tokens in parallel to effectively reduce latency for real-time speech interaction, and can potentially benefit from bidirectional context.
% However, current mainstream speech NAR paradigms still face significant limitations: 
% single-pass methods based on conditional independence weaken long-range 
% dependency modeling and global 
% consistency~\citep{fang2024llama,luo2025openomni}; 
% extra alignment components or supervision increase training 
% complexity~\citep{wang2024maskgct,eskimez2024e2}; 
% and resolving the inherent one-to-many uncertainty in speech generation 
% remains challenging, often requiring additional control signals~\citep{liu2025diffstyletts,xin2024rall}.
While non-autoregressive (NAR) generation offers a promising direction by predicting tokens in parallel to effectively reduce latency for real-time speech interaction and potentially benefiting from bidirectional context, current mainstream speech NAR paradigms still face significant limitations: 
single-pass methods based on conditional independence weaken long-range 
dependency modeling and global 
consistency~\citep{fang2024llama,luo2025openomni}; 
extra alignment components or supervision increase training 
complexity~\citep{wang2024maskgct,eskimez2024e2}; 
and resolving the inherent one-to-many uncertainty in speech generation 
remains challenging, often requiring additional control signals~\citep{liu2025diffstyletts,xin2024rall}. These constraints motivate exploring an alternative NAR paradigm that retains parallel efficiency while addressing quality challenges.

In contrast, Masked Diffusion Modeling (MDM) offers a promising paradigm. 
Proven effective in large language 
models~\citep{nie2025large,zhu2025lladamoe,ye2025dream} and multimodal 
tasks~\citep{yang2025mmada,you2025llada,gong2025diffucoder}, MDM formulates generation as 
forward masking and reverse recovery, iteratively predicting masked tokens under updated contexts, 
forming globally consistent sequences without requiring extra alignment components. 
However, adapting MDM to streaming speech generation introduces two critical challenges.
First, training-inference distribution mismatch: in streaming generation, MDM sequentially generates blocks of tokens by performing diffusion within each block and conditioning on previous blocks~\citep{arriola2025block}. 
% However, standard MDM adopts Global Bernoulli Masking~\citep{nie2025large,zhu2025llada}, causing all blocks in a sample to have similar masking ratios, while block diffusion decoding requires each block to progress from fully masked to fully visible.
However, standard MDM adopts Global Bernoulli Masking~\citep{nie2025large}, making blocks within the same sample appear similarly masked and mismatching block diffusion decoding, where past blocks are fully visible while the current block is progressively revealed from fully masked to fully visible.

%Global Bernoulli masking 的问题在于它对整段序列用同一个概率独立掩码，导致在一个训练样本里切成多个 block 后，各个 block 往往呈现被遮住的比例差不多的均匀状态。可是在 streaming 的 block diffusion decoding 中，推理时的输入状态是高度不均匀的：历史 block 已经完全可见，当前 block 则需要从几乎全遮逐步恢复到完全可见。训练时常见的各块相似掩码与推理时需要的块间差异大、当前块逐步变化不一致，从而造成明显的训练-推理分布不匹配。

Second, cumulative latency from iterative refinement: to maintain generation quality, diffusion decoding still requires a number of denoising steps proportional to the sequence length, resulting in limited inference efficiency~\citep{wu2025fast,chen2025dparallel}.
Existing methods for accelerating iterative refinement also fall short.
KV Cache adaptation improves overall 
throughput~\citep{wu2025fast,ma2025dkv,hu2025accelerating} but cannot reduce first-chunk 
latency; distillation methods require complex multi-model coordination 
and incur high training costs on long speech 
sequences~\citep{kim2025cdlm,zhu2025di,wu2025fastv2}.

Therefore, we explore a new paradigm for speech LLMs using MDM for speech generation, and introduce VocalNet-MDM, an MDM-based speech LLM.
Our approach introduces two mechanisms. First, Hierarchical Block-wise Masking mitigates the training--inference mismatch in block diffusion decoding. Instead of Global Bernoulli Masking, it selects blocks and applies diverse intra-block masking ratios so that training covers the intermediate masked states encountered as each block is gradually denoised. Second, Iterative Self-Distillation reduces diffusion steps by using multi-step block refinement as self-supervision. It transfers more certain prediction distributions from later denoising steps to earlier steps via parallel updates across blocks, 
enabling few-step decoding while preserving multi-step quality.

Experiments demonstrate that VocalNet-MDM, trained on only 6K hours of data, 
achieves 3.7$\times$ to over 10$\times$ decoding speedup compared to the 
AR baseline and reduces first-chunk latency by 34\%, while 
maintaining competitive WER and achieving state-of-the-art text quality 
and speech naturalness.
These results validate MDM as a viable alternative to AR decoding and suggest a new, scalable direction for low-latency, efficient speech LLMs.
Our key contributions are:

\begin{itemize}
    \item \textbf{A new paradigm for speech LLMs via Masked Diffusion Modeling.} We systematically explore MDM in spoken dialogue and empirically validate its compatibility with NAR speech generation.

    \item \textbf{Hierarchical Block-wise Masking for training-inference
    alignment.} By explicitly covering progressive masked states in block diffusion decoding, we mitigate training-inference mismatch and enhance generation quality under streaming inference.

    \item \textbf{Iterative Self-Distillation for low-latency inference.}
    We compress multi-step refinement into few-step decoding via parallel block-wise distillation, retaining multi-step quality without complex multi-model coordination. 
\end{itemize}

%% file: sections/2_Preliminary.tex
\section{Preliminary}
\label{sec:preliminary}

\paragraph{Masked Diffusion Modeling.}
Masked diffusion language models define generation via a forward masking and a reverse recovery~\citep{nie2025large,zhu2025llada}. Given a clean token sequence $\mathbf{s}_{1:T}\in\mathcal{V}^T$ and conditioning information $\mathbf{c}$, the forward process produces a partially masked sequence $\tilde{\mathbf{s}}$ by independently selecting a subset of positions to replace with a special mask token \texttt{[MASK]}, while leaving the remaining tokens unchanged. Let $\mathcal{M}\subseteq\{1,\ldots,T\}$ denote the set of masked positions in $\tilde{\mathbf{s}}$. The reverse process recovers the data by iteratively predicting the masked tokens.

The core component is a mask predictor $p_\theta(\cdot \mid \tilde{\mathbf{s}}, \mathbf{c})$, typically a Transformer without causal masking, which predicts all masked tokens in parallel conditioned on the partially observed sequence. Training minimizes the cross-entropy over masked positions:
\begin{equation}
\mathcal{L}(\theta)
=
\mathbb{E}_{\mathbf{s},\mathbf{c},\mathcal{M}}
\left[
\sum_{t \in \mathcal{M}} -\log p_\theta(s_t \mid \tilde{\mathbf{s}}, \mathbf{c})
\right],
\label{eq:mdm_loss}
\end{equation}
where $\tilde{\mathbf{s}}$ is deterministically obtained from $(\mathbf{s},\mathcal{M})$ by setting $\tilde{s}_t=M$ for $t\in\mathcal{M}$ and $\tilde{s}_t=s_t$ otherwise, and the expectation is taken over training samples $(\mathbf{s},\mathbf{c})$ and masked sets $\mathcal{M}$ drawn from a masking strategy. Unless otherwise specified, a common default is Global Bernoulli Masking, which samples a sequence-level masking ratio and then masks each position independently with this shared probability, spanning corruption levels from lightly masked inputs to highly masked sequences.

\paragraph{Decoding with Block Diffusion.}
In standard MDM decoding, one fixes a target length $T$, initializes $\tilde{\mathbf{s}}^{(0)}=\texttt{[MASK]}^T$, and iteratively predicts and updates masked tokens until all tokens are recovered~\citep{zhu2025llada}.
This whole-sequence denoising requires the full output canvas to be specified upfront, which is incompatible with streaming generation.

Block diffusion instead performs blockwise decoding: it generates blocks sequentially and denoises tokens within each block via masked diffusion~\citep{arriola2025block}.
Partition $\mathbf{s}_{1:T}$ into $K_{\text{blk}}=\lceil T/B\rceil$ contiguous blocks $\mathbf{s}^{(1)},\dots,\mathbf{s}^{(K_{\text{blk}})}$ of size $B$.
The induced model distribution factorizes as
\begin{equation}
\label{eq:prelim_block_factor}
p_\theta(\mathbf{s}_{1:T}\mid \mathbf{c})
=
\prod_{k=1}^{K_{\text{blk}}}
p_\theta\!\left(\mathbf{s}^{(k)} \mid \mathbf{s}^{(<k)}, \mathbf{c}\right).
\end{equation}
Decoding proceeds sequentially for $k=1,\dots,K_{\text{blk}}$: initialize the current block as $\tilde{\mathbf{s}}^{(k,0)}=M^{B}$, then simulate the reverse denoising process restricted to this block by iteratively applying the mask predictor conditioned on $\mathbf{s}^{(<k)}$ and $\mathbf{c}$, yielding a sample $\mathbf{s}^{(k)} \sim p_\theta(\mathbf{s}^{(k)} \mid \mathbf{s}^{(<k)}, \mathbf{c})$, which is appended to the prefix.
Varying the block size interpolates between autoregressive decoding at $B=1$ and whole-sequence diffusion when $K_{\text{blk}}=1$.

%% file: sections/3_Methodology.tex
\section{Methodology}
\label{sec:method}

\begin{figure*}[t]
  \centering
  \includegraphics[width=1\textwidth]{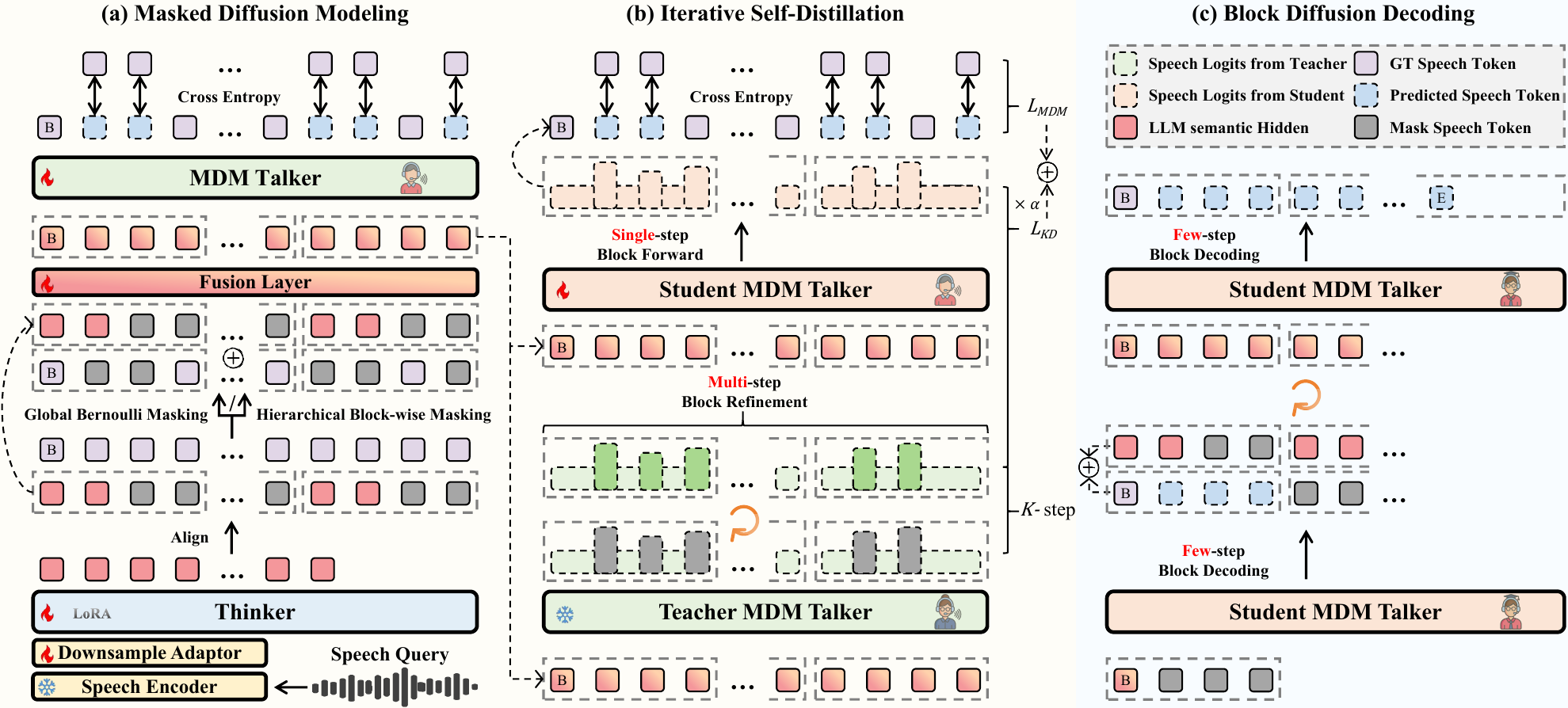}
  \caption{
    \textbf{Overview of our approach.}
    (a) Masked Diffusion Modeling: a Thinker--Talker speech LLM architecture where speech generation is performed via MDM.
    (b) Iterative Self-Distillation: transferring more certain distributions from later to earlier denoising steps to enable few-step inference.
    (c) Block Diffusion Decoding: streaming inference that denoises tokens within fixed-size blocks over diffusion steps.
  }
  \label{fig:architecture}
\end{figure*}

\subsection{Overall Model Architecture}
\label{sec:method_architecture}

% We develop a Thinker--Talker speech LLMs architecture derived from Qwen3-Omni~\citep{Qwen3-Omni} (Figure~\ref{fig:architecture}).

We develop VocalNet-MDM, a speech LLM that adopts the Thinker--Talker paradigm introduced by Qwen3-Omni~\citep{Qwen3-Omni}, as illustrated in Figure~\ref{fig:architecture}.

Given an input waveform $\mathbf{x}$, a Speech Encoder followed by a Downsample Adaptor produces a sequence of continuous representations $\mathbf{r}_{1:L}$. Conditioned on $\mathbf{r}_{1:L}$, the Thinker autoregressively generates the textual tokens $\mathbf{y}_{1:N}$ and outputs the corresponding hidden states $\mathbf{h}_{1:N}$:
\begin{equation}
\mathbf{y}_{1:N}, \mathbf{h}_{1:N} = \text{Thinker}(\mathbf{r}_{1:L}).
\end{equation}

% The Talker generates a discrete speech token sequence $\mathbf{s}_{1:T}$ of length $T$. Since typically $T \gg N$, we apply intra-block sparse semantic anchor alignment to construct $\mathbf{h}'_{1:T}$ from $\mathbf{h}_{1:N}$ under low-latency streaming constraints. 
The Talker generates a discrete speech token sequence $\mathbf{s}_{1:T}$ of length $T$. Since typically $T \gg N$, we apply intra-block sparse semantic anchor alignment to construct $\mathbf{h}'_{1:T}$ from $\mathbf{h}_{1:N}$.
Specifically, we partition the speech timeline into blocks of size $B$ and place semantic anchors at the first $Q$ positions of each block, setting all other positions to $\mathbf{0}$. Anchor semantics are assigned in an order-preserving, prefix-only manner to avoid future semantic leakage. Details are provided in Appendix~\ref{sec:appendix_anchor_properties}.

During training, the Talker processes a corrupted input sequence $\tilde{\mathbf{s}}_{1:T}$, obtained by masking $\mathbf{s}_{1:T}$ with \texttt{[MASK]} token. Its embedding sequence is denoted as $\mathbf{E}^s_{1:T}=\text{Embed}(\tilde{\mathbf{s}}_{1:T})$. We add the speech embeddings to the aligned semantics and apply a two-layer feed-forward Fusion module:
\begin{equation}
\begin{aligned}
\mathbf{u}_{1:T} &= \text{ReLU}\!\left(\mathbf{W}_1(\mathbf{E}^s_{1:T}+\mathbf{h}'_{1:T})+\mathbf{b}_1\right),\\
\mathbf{e}_{1:T} &= \mathbf{W}_2\mathbf{u}_{1:T}+\mathbf{b}_2.
\end{aligned}
\end{equation}
The fused features $\mathbf{e}_{1:T}$ are fed into the Talker, which performs MDM to conditionally predict the masked speech tokens. Finally, the predicted discrete speech tokens are converted into waveforms using the pretrained flow matching model and the HiFi-GAN vocoder from CosyVoice2~\citep{du2024cosyvoice}, enabling high fidelity and low latency streaming speech reconstruction.

\subsection{Masked Diffusion Modeling}
\label{sec:method_mdm}
Talker adopts the non-autoregressive MDM paradigm (Figure~\ref{fig:architecture}(a)). Given the fused feature sequence $\mathbf{e}_{1:T}$, the Talker predicts masked positions in parallel under a conditional independence approximation, instantiating an intra-block parallel, inter-block causal modeling scheme.

\paragraph{Block-Causal Attention Mechanism.}
To enable streaming generation, we adopt Block-Causal attention~\citep{arriola2025block}. We divide a length-$T$ sequence into consecutive blocks of size $B$. Tokens within the same block are fully visible to each other, enabling parallel computation within the block. Across blocks, attention is restricted causally: each token can attend to all positions in the current block and all preceding blocks, ensuring strict inter-block causality.

\begin{figure}[t]
  \centering
  \includegraphics[width=0.48\textwidth]{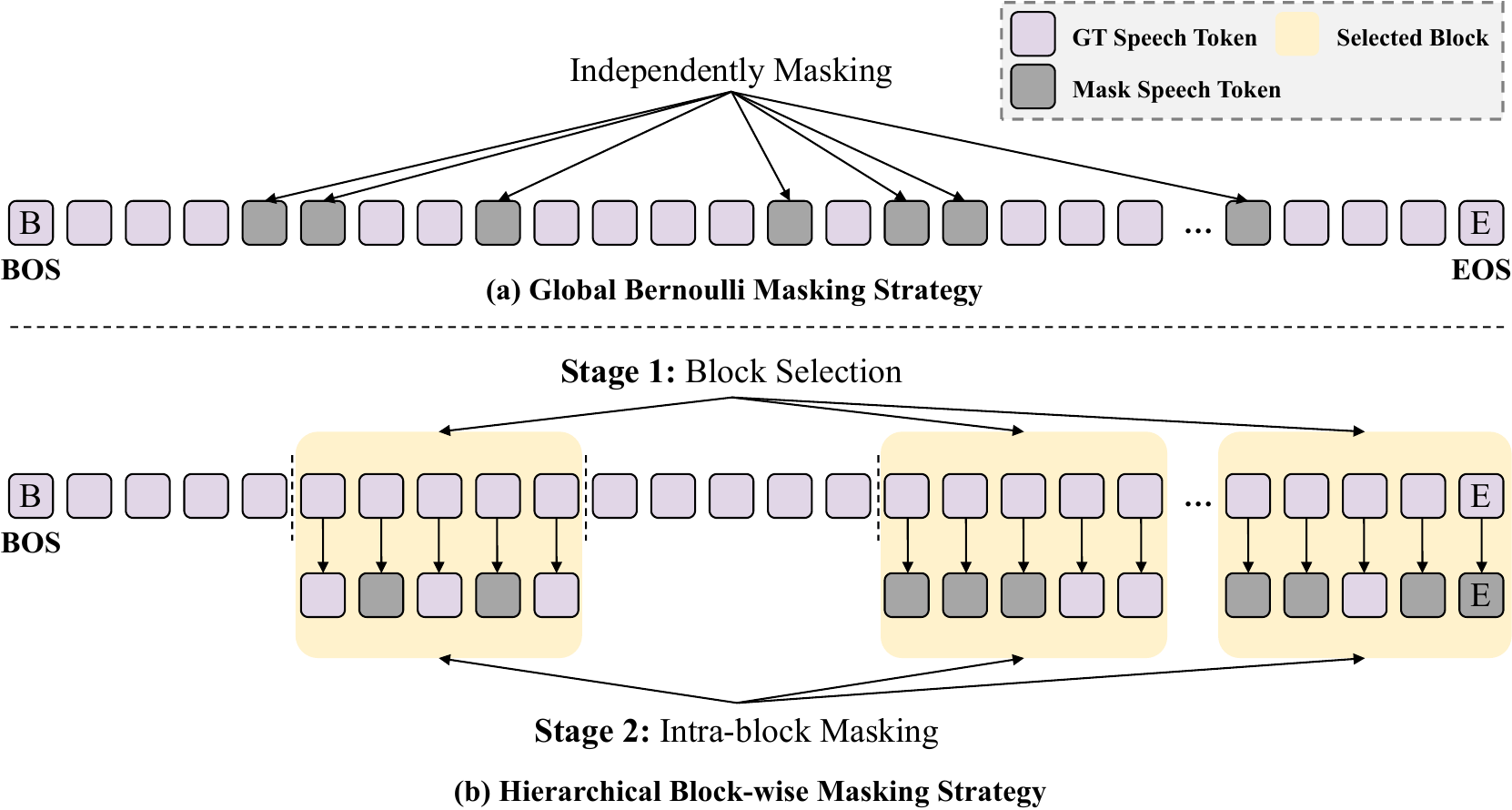}
    \caption{
      \textbf{Masking strategies.}
      (a) Global Bernoulli Masking: each token masked independently 
      with uniform probability.
      (b) Hierarchical Block-wise Masking: block selection followed 
      by intra-block token masking.
    }
  \label{fig:masking_strategy}
\end{figure}

\paragraph{Masking Strategies.}
To support MDM training, we adopt Global Bernoulli Masking~\citep{nie2025large} where each token is independently masked with probability $\gamma_g \sim U(\gamma_g^{\min},\gamma_g^{\max})$, providing uniform training signals across all positions. However, to better align training with the intermediate patterns encountered during block diffusion decoding, we propose a Hierarchical Block-wise Masking strategy. By hierarchically sampling masking ratios, this approach explicitly covers the progressive states where blocks transition from fully masked to fully visible, thereby bridging the training-inference distribution gap. Figure~\ref{fig:masking_strategy} illustrates the two masking strategies. 

Concretely, for each sample we partition the valid speech token indices into $K_{\text{blk}}=\lceil T/B\rceil$ consecutive blocks, where the index set of the $k$-th block is
\begin{equation}
\begin{aligned}
\mathcal{I}_k &= \{t \mid (k-1)B < t \le \min(kB, T)\}, \\
k &= 1, \dots, K_{\text{blk}}.
\end{aligned}
\label{eq:block_partition}
\end{equation}
Masking is sampled via a two-stage randomized process: we first select a random subset of blocks $\mathcal{K}_{\text{mask}}$ based on a block-level ratio $\gamma_c \sim U(\gamma_c^{\min},\gamma_c^{\max})$, then within each selected block we mask a fraction of positions determined by $\gamma_t \sim U(\gamma_t^{\min},\gamma_t^{\max})$ to form $\mathcal{M}$. The complete algorithmic procedure is provided in Appendix~\ref{sec:appendix_masking}.

\paragraph{MDM Training Objective.}
Given the masked set $\mathcal{M}$, the corrupted input $\tilde{\mathbf{s}}_{1:T}$, and the aligned semantic stream $\mathbf{h}'_{1:T}$, we minimize the cross-entropy loss on masked positions:
\begin{equation}
\mathcal{L}_{\text{MDM}} = - \sum_{t \in \mathcal{M}} \log P\!\left(s_t \mid \tilde{\mathbf{s}}_{1:T}, \mathbf{h}'_{1:T}\right).
\end{equation}

\subsection{Accelerated Inference via Iterative Self-Distillation}
\label{sec:method_distillation}
Although MDM alleviates the serial bottleneck of autoregressive decoding via parallel prediction, high-quality speech typically relies on multi-step iterative unmasking, which incurs additional latency. To balance quality and latency, we propose Iterative Self-Distillation (Figure~\ref{fig:architecture}(b)),
% compressing the progressive disambiguation capability of multi-step refinement into a student model for comparable quality with substantially reduced inference overhead.
distilling the teacher’s more confident predictions from later denoising steps into earlier ones, enabling few-step inference while preserving multi-step quality.

During training, the Teacher is a frozen replica of the model, and the Student is optimized with distillation targets. The Teacher runs $K$ iterations of MDM and dynamically collects supervision signals to construct a target tensor $\mathbf{Z}_{\text{tea}}$.

\paragraph{Block-wise Teacher Update.}
Given the corrupted training input $\tilde{\mathbf{s}}^{(0)}_{1:T}$ and its masked set $\mathcal{M}$, at iteration $j$ ($1\le j\le K$) the Teacher produces logits $\mathbf{z}^{(j)}_{\text{tea}}(t)$ conditioned on the current sequence $\tilde{\mathbf{s}}^{(j-1)}_{1:T}$ and the semantic stream $\mathbf{h}'_{1:T}$. We adopt a block-wise update strategy with parallel updates across blocks: within each block of size $B$, we independently reveal a subset of masked positions based on confidence. This avoids inefficient global ranking and block-by-block serial updates. For position $t$, we define the confidence as
\begin{equation}
\label{eq:confidence}
u^{(j)}_t = \max_{v\in\mathcal{V}} \text{Softmax}\!\left(\mathbf{z}^{(j)}_{\text{tea}}(t)\right)_v.
\end{equation}
For each block, at iteration $j$ we select the top-$n_j$ masked positions with relatively highest confidence within the block and replace their \texttt{[MASK]} tokens with argmax predictions, yielding $\tilde{\mathbf{s}}^{(j)}$.

\paragraph{Update Scheduling and Dynamic Logit Collection.}
Let $R_j$ denote the number of remaining masked positions in a given block at the beginning of iteration $j$. The number of updates $n_j$ is determined by an even-allocation schedule:
\begin{equation}
\label{eq:even_schedule}
r_j = K-j+1, \quad
n_j =
\begin{cases}
0, & R_j = 0,\\
\left\lceil \dfrac{R_j}{r_j} \right\rceil, & \text{otherwise}.
\end{cases}
\end{equation}
When a position $t$ is revealed at iteration $j$, we record the Teacher logits at that moment as its distillation target:
\begin{equation}
\mathbf{Z}_{\text{tea}}(t) = \mathbf{z}^{(j)}_{\text{tea}}(t), \quad \forall t \in \mathcal{U}^{(j)},
\end{equation}
where $\mathcal{U}^{(j)}$ denotes the set of positions revealed at iteration $j$.

\paragraph{Distillation Objective.}
The Student produces logits $\mathbf{z}_{\text{stu}}(t)$ in a single forward pass and matches the Teacher target tensor $\mathbf{Z}_{\text{tea}}$. We adopt Reverse KL with temperature $\tau$ as the distillation loss:
\begin{equation}
\begin{split}
\mathcal{L}_{\text{KD}}
&= \frac{\tau^2}{|\mathcal{M}|}\sum_{t\in\mathcal{M}}
\mathrm{KL}\Big(
\text{Softmax}(\mathbf{z}_{\text{stu}}(t)/\tau) \\
&\qquad\qquad\Big\|\,
\text{Softmax}(\mathbf{Z}_{\text{tea}}(t)/\tau)
\Big).
\end{split}
\end{equation}
The final objective is a weighted combination of the distillation loss and the original masked cross-entropy:
\begin{equation}
\mathcal{L}_{\text{Distill}} = \alpha \mathcal{L}_{\text{KD}} + (1-\alpha)\mathcal{L}_{\text{MDM}}.
\end{equation}

\paragraph{Training Strategy.}
The training proceeds in three progressive stages. We first align 
the Thinker to the audio modality using speech-to-text data via 
LoRA. We then train the Talker with $\mathcal{L}_{\text{MDM}}$ 
using Global Bernoulli masking to establish parallel prediction. 
Finally, we fine-tune the Talker with $\mathcal{L}_{\text{Distill}}$ 
using Hierarchical Block-wise Masking.

\subsection{Block Diffusion Decoding}
\label{sec:method_decoding}
To enable low-latency interaction, we generate speech in consecutive
fixed-size blocks of length $B$ using Block Diffusion Decoding
(Figure~\ref{fig:architecture}(c)). Given a prefix $\mathbf{s}_{<t_0}$, we
initialize the the subsequent block with \texttt{[MASK]} and denoise it for $K$ diffusion
steps. At each step $j$, the Talker predicts distributions for masked positions
conditioned on $\mathbf{h}'$ and $\mathbf{s}_{<t_0}$, computes confidence via
Eq.~\eqref{eq:confidence}, and unmasks the top-$n_j$ tokens following
Eq.~\eqref{eq:even_schedule}. The resulting block is then appended to the prefix for the next iteration; if an EOS token is generated, we truncate the sequence at the earliest EOS and terminate generation.

%% file: sections/4_Experiments.tex
\section{Experiments}

\subsection{Experimental Setup}

\paragraph{Datasets.}
We follow the data construction protocol of VocalNet~\citep{wang2025vocalnet1}, combining VoiceAssistant-400K from Mini-Omni and UltraChat from SLAM-Omni~\citep{xie2024mini,chen2025slam}. VoiceAssistant-400K is generated by GPT-4o, yielding approximately 430K single-turn query-response pairs. UltraChat is split from multi-round dialogues into single-turn interactions, producing roughly 300K samples. Speech responses are synthesized via CosyVoice2-0.5B~\citep{du2024cosyvoice}, and we use its corresponding discrete codec tokens for training and decoding. The training set totals around 730K examples, corresponding to approximately 6k hours of speech. 

\begin{table*}[t]
    \centering
    \resizebox{\linewidth}{!}{
        \begin{tabular}{lccccccccc}
            \toprule
            \multirow{2}{*}{\textbf{Model}} & \multicolumn{2}{c}{\textbf{Efficiency}} & \multicolumn{5}{c}{\textbf{Text Quality}} & \multicolumn{2}{c}{\textbf{Speech Quality}} \\
            \cmidrule(lr){2-3} \cmidrule(lr){4-8} \cmidrule(lr){9-10}
            & \textbf{TPS}$\uparrow$ & \textbf{RTF}$\downarrow$ & \textbf{AlpacaEval} & \textbf{Llama Q.} & \textbf{TriviaQA} & \textbf{Web Q.} & \textbf{Avg} & \textbf{WER}$\downarrow$ & \textbf{UTMOS}$\uparrow$ \\
            \midrule
            SLAM-Omni \citep{chen2025slam}      & 61.13   & 0.8180  & 3.50 & 2.94 & 0.39 & 0.84 & 1.92 & 5.78  & 4.46 \\
            VITA-Audio \citep{long2025vita}     & 64.62   & 0.1934 & 6.93 & 7.43 & 4.27 & 5.23 & 5.96 & 3.83  & 4.26 \\
            GLM-4-Voice \citep{zeng2024glm}     & 39.06   & 0.3200 & 5.86 & 7.74 & 4.95 & 5.56 & 6.03 & 11.90 & 4.23 \\
            MiniCPM-o \citep{MiniCPM-o-2.6}     & 212.16  & 0.2205 & 6.13 & 7.72 & \textbf{6.43} & \textbf{7.16} & 6.86 & 9.52  & 4.14 \\
            Qwen2.5-Omni \citep{xu2025qwen2}    & 58.82   & 0.8484 & 6.01 & 7.90 & 5.89 & 6.88 & 6.67 & \textbf{2.31} & 4.34 \\
            Kimi-Audio \citep{ding2025kimi}     & 30.77   & 0.4062 & 6.49 & 8.10 & 6.15 & 7.10 & 6.96 & 14.71 & 2.87 \\
            VocalNet-8B \citep{wang2025vocalnet1}                         & 374.81  & 0.0667 & 7.12 & 7.95 & 6.24 & 6.48 & 6.95 & 3.64  & \textbf{4.49} \\
            MiMo-Audio \citep{coreteam2025mimoaudio} & 476.51  & 0.4197 & 7.35 & 7.70 & 4.90 & 5.80 & 6.44 & 6.13  & 3.68 \\
            \midrule
            Baseline-AR (NTP) & 204.64  & 0.1222 & \textbf{7.43} & \textbf{8.27} & 6.15 & 6.42 & \textbf{7.07} & 10.66 & 4.48 \\
            Baseline-AR (MTP) & 374.81  & 0.0667 & \textcolor{gray}{\textbf{7.43}} & \textcolor{gray}{\textbf{8.27}} & \textcolor{gray}{6.15} & \textcolor{gray}{6.42} & \textcolor{gray}{\textbf{7.07}} & 4.05  & \textbf{4.49} \\
            \cellcolor{gray!15}VocalNet-MDM (Step 16) & \cellcolor{gray!15}221.58  & \cellcolor{gray!15}0.1128 & \cellcolor{gray!15}\textbf{7.43} & \cellcolor{gray!15}\textbf{8.27} & \cellcolor{gray!15}6.15 & \cellcolor{gray!15}6.42 & \cellcolor{gray!15}\textbf{7.07} & \cellcolor{gray!15}5.53 & \cellcolor{gray!15}\textbf{4.49} \\
            \cellcolor{gray!15}VocalNet-MDM (Step 8) & \cellcolor{gray!15}390.72  & \cellcolor{gray!15}0.0639 & \cellcolor{gray!15}\textcolor{gray!60}{\textbf{7.43}} & \cellcolor{gray!15}\textcolor{gray!60}{\textbf{8.27}} & \cellcolor{gray!15}\textcolor{gray!60}{6.15} & \cellcolor{gray!15}\textcolor{gray!60}{6.42} & \cellcolor{gray!15}\textcolor{gray!60}{\textbf{7.07}} & \cellcolor{gray!15}5.55 & \cellcolor{gray!15}\textbf{4.49} \\
            \cellcolor{gray!15}VocalNet-MDM (Step 4)  & \cellcolor{gray!15}757.58  & \cellcolor{gray!15}0.0329 & \cellcolor{gray!15}\textcolor{gray!60}{\textbf{7.43}} & \cellcolor{gray!15}\textcolor{gray!60}{\textbf{8.27}} & \cellcolor{gray!15}\textcolor{gray!60}{6.15} & \cellcolor{gray!15}\textcolor{gray!60}{6.42} & \cellcolor{gray!15}\textcolor{gray!60}{\textbf{7.07}} & \cellcolor{gray!15}5.34 & \cellcolor{gray!15}\textbf{4.49} \\
            \cellcolor{gray!15}VocalNet-MDM (Step 2)  & \cellcolor{gray!15}1327.80 & \cellcolor{gray!15}0.0188 & \cellcolor{gray!15}\textcolor{gray!60}{\textbf{7.43}} & \cellcolor{gray!15}\textcolor{gray!60}{\textbf{8.27}} & \cellcolor{gray!15}\textcolor{gray!60}{6.15} & \cellcolor{gray!15}\textcolor{gray!60}{6.42} & \cellcolor{gray!15}\textcolor{gray!60}{\textbf{7.07}} & \cellcolor{gray!15}6.10 & \cellcolor{gray!15}4.47 \\
            \cellcolor{gray!15}VocalNet-MDM (Step 1)  & \cellcolor{gray!15}\textbf{2153.43} & \cellcolor{gray!15}\textbf{0.0116} & \cellcolor{gray!15}\textcolor{gray!60}{\textbf{7.43}} & \cellcolor{gray!15}\textcolor{gray!60}{\textbf{8.27}} & \cellcolor{gray!15}\textcolor{gray!60}{6.15} & \cellcolor{gray!15}\textcolor{gray!60}{6.42} & \cellcolor{gray!15}\textcolor{gray!60}{\textbf{7.07}} & \cellcolor{gray!15}6.23 & \cellcolor{gray!15}4.46 \\
            \bottomrule
        \end{tabular}
    }
    \caption{Comparison of VocalNet-MDM with open-source speech LLMs and an autoregressive baseline (Baseline-AR). Baseline-AR uses the same backbone as VocalNet-MDM and differs only in the talker, using next-token prediction (NTP) or multi-token prediction (MTP), whereas VocalNet-MDM uses masked diffusion modeling (MDM). \textbf{Bold} indicates the best performance. \textcolor{gray}{Gray} entries denote metrics shared across variants.} 
    \label{tab:main_results}
\end{table*}

% \paragraph{Model Configuration.}
% We use Whisper-large-v3~\citep{radford2023robust} with 5$\times$ downsampling as the speech encoder, Qwen3-8B~\citep{yang2025qwen3} as the thinker, and four Qwen3-style Transformer layers as the talker. The AR baselines share the same architecture and train the talker autoregressively with next-token prediction (NTP) and multi-token prediction (MTP), where the MTP setting follows \citet{wang2025vocalnet1}.
\paragraph{Model Configuration.}
We use Whisper-large-v3~\citep{radford2023robust} with 5$\times$ downsampling as the speech encoder. The Thinker is initialized from Qwen3-8B~\citep{yang2025qwen3}. The Talker shares a similar architectural design with the Thinker but uses 4 Transformer layers and is trained separately from scratch. The AR baselines share the same architecture and train the Talker autoregressively with next-token prediction (NTP) and multi-token prediction (MTP), where MTP follows \citet{wang2025vocalnet1}.

\paragraph{Training Details.}
We set the block size, distillation iterations, and semantic anchors to $B=16$, $K=4$, and $Q=4$, respectively. Global Bernoulli masking probability is sampled from $[\gamma_g^{\min},\gamma_g^{\max}]=[0.3,0.8]$. For Hierarchical Block-wise Masking, block-level ratio is sampled from $[\gamma_c^{\min},\gamma_c^{\max}]=[0.5,1.0]$, and intra-block ratio from $[\gamma_t^{\min},\gamma_t^{\max}]=[0.3,1.0]$. We use temperature $\tau=2.0$ and weight $\alpha=0.7$. Distillation uses a frozen teacher, adding only forward refinements.

We use AdamW with learning rate $2\times 10^{-4}$ and batch size 32 on A100 GPUs. We train the Thinker and downsampling adapter for 1 epoch, the Talker for 8 epochs with $\mathcal{L}_{\text{MDM}}$ using Global Bernoulli Masking. For distillation fine-tuning, we create a frozen copy of this checkpoint as the teacher and continue training the Talker for 4 epochs with $\mathcal{L}_{\text{Distill}}$ using Hierarchical Block-wise Masking. The frozen teacher performs only forward refinements without parameter updates.

\paragraph{Evaluation.}
We evaluate on the English subsets of OpenAudioBench~\citep{li2025baichuan}: AlpacaEval~\citep{alpaca_eval}, Llama Questions~\citep{nachmani2023spoken}, TriviaQA~\citep{joshi2017triviaqa}, and Web Questions~\citep{berant2013semantic}. Text quality is scored by Qwen-max for correctness and relevance on a 0--10 scale. See Appendix~\ref{sec:appendix_eval} for details. Speech quality is measured by UTMOS~\citep{saeki2022utmos} for naturalness and WER using Whisper-large-v3 transcriptions for speech--text alignment.

We report efficiency using TPS and RTF, defined as generated speech tokens divided by inference latency, and inference latency divided by speech duration, respectively. Latency is measured for speech token generation only. For aligned multimodal models, we measure Talker latency; for native multimodal models, we measure speech-token generation latency in the unified model. Token counting includes only speech tokens. All measurements are conducted on a single L20 GPU. 

\subsection{Main Results}

Table~\ref{tab:main_results} compares VocalNet-MDM with open-source speech LLMs and an autoregressive baseline (Baseline-AR) in terms of efficiency, text quality, and speech quality. For VocalNet-MDM, Step~$K$ denotes $K$ diffusion steps per block.

\paragraph{Text Quality.} VocalNet-MDM achieves the best average text quality score of 7.07. Specifically, it obtains 7.43 on AlpacaEval and 8.27 on Llama Questions, both representing the highest scores across all models, demonstrating that our approach preserves the strong semantic understanding and reasoning capabilities of the Qwen3-8B base model. On TriviaQA and Web Questions, VocalNet-MDM achieves 6.15 and 6.42 respectively, ranking third. Overall, VocalNet-MDM demonstrates balanced performance across all four evaluation datasets, achieving the best comprehensive text quality.

\paragraph{Efficiency Advantages.} 
VocalNet-MDM demonstrates substantial efficiency advantages over both AR baselines and open-source models. Relative to Baseline-AR (NTP), Step~4 improves TPS by 3.7$\times$ and reduces RTF by 73\%, while Step~1 achieves even more dramatic gains with 10.5$\times$ higher TPS and 90.5\% RTF reduction. Relative to Baseline-AR (MTP), Step~4 achieves 2.0$\times$ higher TPS, and Step~1 delivers 5.7$\times$ speedup. Among prior models, Step~4 attains 1.6$\times$ higher TPS than the second-fastest model, MiMo-Audio, while Step~1 achieves 4.5$\times$ higher TPS. The RTF of Step~4 and Step~1 are also markedly lower than most open-source models, which typically exceed 0.4.

\paragraph{Speech Quality.} 
VocalNet-MDM maintains competitive speech quality across configurations. Step~16, 8, 4, 2, and 1 all achieve UTMOS scores of 4.49, 4.49, 4.49, 4.47, and 4.46 respectively, with Step~4 through 16 exceeding all other models in naturalness. For speech--text alignment, Step~1 through 4 achieve WER ranging from 5.34 to 6.23, substantially improving over Baseline-AR (NTP) at 10.66 and approaching Baseline-AR (MTP) at 4.05. Compared to other models, Step~4 outperforms SLAM-Omni (5.78), MiMo-Audio (6.13), MiniCPM-o (9.52), GLM-4-Voice (11.90), and Kimi-Audio (14.71), while Step~1 maintains competitive performance with WER of 6.23.

\paragraph{Efficiency--Quality Trade-off.}
VocalNet-MDM offers flexible operating points to balance efficiency and quality. Step~4 serves as a strong default: it matches Step~16 in naturalness with an identical UTMOS of 4.49, maintains a low WER of 5.34, and incurs lower training cost. We therefore adopt $K=4$ as the default distillation iteration count. Step~1 targets latency-critical settings, achieving the highest throughput of 2153.43 TPS and lowest RTF of 0.0116 while preserving competitive quality with WER 6.23 and UTMOS 4.46.

\begin{table}[t]
    \centering
    \resizebox{\columnwidth}{!}{%
    \begin{tabular}{c|c}
        \toprule
        \textbf{Model} & \textbf{First chunk latency (ms)}$\downarrow$ \\
        \midrule
        SLAM-Omni       & 742.32 $\pm$ 38.22 \\
        VITA-Audio      & 512.64 $\pm$ 33.55 \\
        GLM-4-Voice     & 1066.02 $\pm$ 21.78 \\
        MiniCPM-o       & 1329.52 $\pm$ 257.27 \\
        Qwen2.5-Omni    & N/A \\
        Kimi-Audio      & 1371.48 $\pm$ 137.50 \\
        VocalNet-8B     & 462.32 $\pm$ 45.07 \\ 
        MiMo-Audio      & N/A \\
        \midrule
        Baseline-AR (NTP) & 555.86 $\pm$ 17.88 \\
        Baseline-AR (MTP) & 481.22 $\pm$ 19.58 \\
        \cellcolor{gray!15}VocalNet-MDM (Step 1)  & \cellcolor{gray!15}\textbf{368.67 $\pm$ 13.82} \\
        \cellcolor{gray!15}VocalNet-MDM (Step 2)  & \cellcolor{gray!15}373.29 $\pm$ 14.00 \\
        \cellcolor{gray!15}VocalNet-MDM (Step 4)  & \cellcolor{gray!15}382.36 $\pm$ 14.34 \\
        \cellcolor{gray!15}VocalNet-MDM (Step 8)  & \cellcolor{gray!15}402.19 $\pm$ 15.08 \\
        \cellcolor{gray!15}VocalNet-MDM (Step 16) & \cellcolor{gray!15}427.45 $\pm$ 16.25 \\
        \bottomrule
    \end{tabular}%
    }
    \caption{First-chunk latency reported as mean $\pm$ standard deviation. N/A indicates models without official streaming inference implementations, for which first-chunk latency cannot be measured.}
    \label{tab:first_chunk_latency}
\end{table}

\subsection{Detailed Efficiency Analysis}
\label{subsec:detailed_efficiency_analysis}

We evaluate first-chunk latency and its components in streaming generation scenarios on a single L20 GPU. For models lacking official streaming implementations, or with adjustable parameters, we standardize the speech chunk length to approximately 0.6 seconds to ensure fair comparison. For models with officially recommended streaming configurations, we adhere to their prescribed settings.

Table~\ref{tab:first_chunk_latency} presents first-chunk latency comparison. VocalNet-MDM achieves the lowest latency across all configurations. Step 1 attains the optimal latency of 368.67ms. Compared to higher-latency models where GLM-4-Voice, MiniCPM-o, and Kimi-Audio exhibit 1066.02ms, 1329.52ms, and 1371.48ms respectively, our Step 16 achieves only 427.45ms, delivering over 2$\times$ reduction.

\begin{figure}[t]
  \centering
  \includegraphics[width=0.48\textwidth]{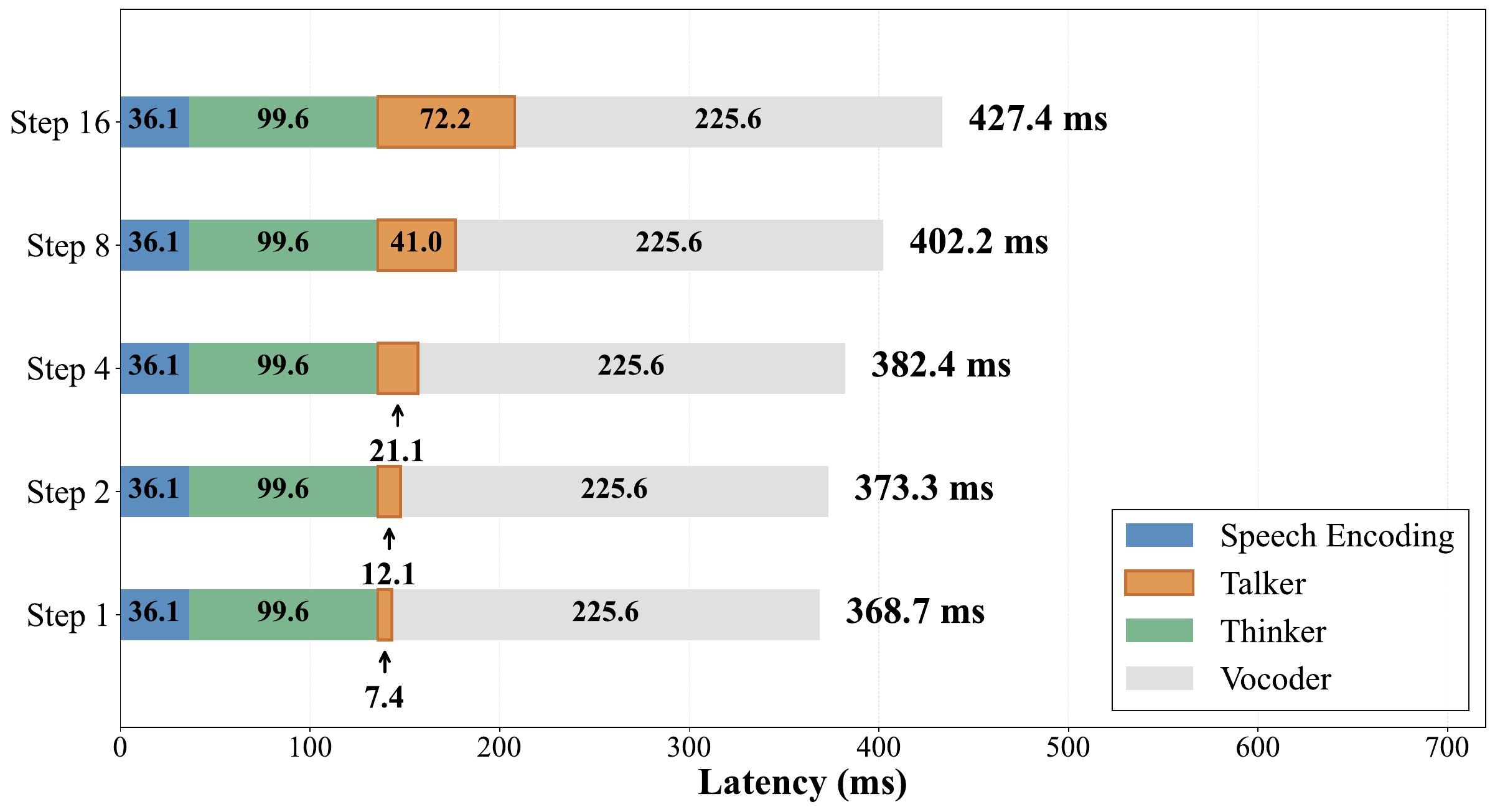}
  \caption{Latency breakdown for generating the first-chunk in VocalNet-MDM at different diffusion steps.}
  \label{fig:figure1}
\end{figure}

Figure~\ref{fig:figure1} further illustrates the latency breakdown of VocalNet-MDM. All configurations maintain consistent latency across Speech Encoding, Thinker, and Vocoder modules at 36.1ms, 99.6ms, and 225.6ms respectively. The Talker module latency decreases with fewer diffusion steps, from 72.2ms at Step 16 to 7.4ms at Step 1. Vocoder latency is dominated by flow matching, which is not essential to the vocoder architecture~\citep{coreteam2025mimoaudio,wang2025vocalnet}. We therefore leave vocoder optimization out of scope.

\begin{table*}[t]
    \centering
    \resizebox{\linewidth}{!}{%
    \begin{tabular}{ll|cc|cc|cc}
        \toprule
        \multirow{2}{*}{\textbf{MDM Training Masking Strategy}} &
        \multirow{2}{*}{\textbf{Distillation FT Masking Strategy}} &
        \multicolumn{2}{c|}{\textbf{Step 4}} &
        \multicolumn{2}{c|}{\textbf{Step 2}} &
        \multicolumn{2}{c}{\textbf{Step 1}} \\
        \cmidrule(lr){3-4} \cmidrule(lr){5-6} \cmidrule(lr){7-8}
        & & \textbf{WER}$\downarrow$ & \textbf{UTMOS}$\uparrow$ & \textbf{WER}$\downarrow$ & \textbf{UTMOS}$\uparrow$ & \textbf{WER}$\downarrow$ & \textbf{UTMOS}$\uparrow$ \\
        \midrule
        Hierarchical Block-wise Masking & N/A & 10.43 & 4.46 & 15.38 & 4.41 & 50.70 & 3.84 \\
        Global Bernoulli Masking & N/A & 6.75 & 4.47 & 9.95 & 4.42 & 39.00 & 3.87 \\
        \midrule
        Global Bernoulli Masking & Global Bernoulli Masking & 7.65 & 4.46 & 7.91 & 4.45 & 9.69 & 4.42 \\
        Global Bernoulli Masking & Hierarchical Block-wise Masking & \textbf{5.34} & \textbf{4.49} & \textbf{6.10} & \textbf{4.47} & \textbf{6.23} & \textbf{4.46} \\
        \bottomrule
    \end{tabular}%
    }
    \caption{Ablation study on masking strategies. MDM Training Masking Strategy denotes the masking strategy used to train the Talker with $\mathcal{L}_{\text{MDM}}$. Distillation FT Masking Strategy denotes the masking strategy used during self-distillation fine-tuning with $\mathcal{L}_{\text{Distill}}$, and N/A indicates that the distillation fine-tuning stage is omitted. The best results are highlighted in \textbf{bold}.}
    \label{tab:masking_ablation}
\end{table*}

\subsection{Impact of Masking Strategy}
We compare masking strategies with identical distillation configurations $\alpha=0.7$ and $\tau=2.0$ (Table~\ref{tab:masking_ablation}). Without distillation, Hierarchical Block-wise Masking exhibits higher WER of 10.43, 15.38, and 50.70 at Step 4, 2, and 1, while Global Bernoulli masking achieves lower WER of 6.75, 9.95, and 39.00. This shows Hierarchical Block-wise Masking struggles to establish parallel prediction capabilities without iterative refinement guidance, while Global Bernoulli masking provides uniform training signals suitable for this purpose.

With Iterative Self-Distillation, switching to Hierarchical Block-wise Masking during fine-tuning reduces WER from 7.65 to 5.34 at Step 4 and from 9.69 to 6.23 at Step 1, representing 30\% and 36\% relative improvements respectively, while maintaining UTMOS scores of 4.49 and 4.46. Hierarchical Block-wise Masking better simulates the intermediate states during block diffusion decoding, enabling more effective compression of multi-step refinement into fewer inference steps. This validates our two-stage training pipeline: first establishing parallel prediction capabilities with Global Bernoulli masking, then switching to Hierarchical Block-wise Masking to align training states with block diffusion decoding patterns.

\begin{table}[t]
    \centering
    \resizebox{\columnwidth}{!}{%
    \begin{tabular}{cc|cc|cc|cc}
        \toprule
        \multirow{2}{*}{$\boldsymbol{\alpha}$} & \multirow{2}{*}{$\boldsymbol{\tau}$} & \multicolumn{2}{c|}{\textbf{Step 4}} & \multicolumn{2}{c|}{\textbf{Step 2}} & \multicolumn{2}{c}{\textbf{Step 1}} \\
        \cmidrule(lr){3-4} \cmidrule(lr){5-6} \cmidrule(lr){7-8}
        & & \textbf{WER}$\downarrow$ & \textbf{UTMOS}$\uparrow$ & \textbf{WER}$\downarrow$ & \textbf{UTMOS}$\uparrow$ & \textbf{WER}$\downarrow$ & \textbf{UTMOS}$\uparrow$ \\
        \midrule
        N/A & N/A & 6.75 & 4.47 & 9.95 & 4.42 & 39.00 & 3.87 \\
        \midrule
        0.7 & 2.0 & \textbf{5.34} & \textbf{4.49} & \textbf{6.10} & \textbf{4.47} & \textbf{6.23} & \textbf{4.46} \\
        0.0 & N/A & 6.29 & 4.47 & 9.77 & 4.43 & 41.27 & 3.86 \\
        0.9 & 2.0 & 10.07 & 4.46 & 10.03 & 4.44 & 15.95 & 4.32 \\
        0.7 & 1.0 & 5.95 & 4.48 & 6.55 & \textbf{4.47} & 9.55 & 4.42 \\
        0.7 & 3.0 & 5.73 & \textbf{4.49} & 6.20 & 4.46 & 9.26 & 4.41 \\
        \bottomrule
    \end{tabular}%
    }
    \caption{Performance under different self-distillation hyperparameter settings. The row with N/A indicates the model trained without the distillation fine-tuning stage. The best results are highlighted in \textbf{bold}.}
    \label{tab:distillation_hyperparams}
\end{table}

\begin{figure}[t]
  \centering
  \includegraphics[width=0.48\textwidth]{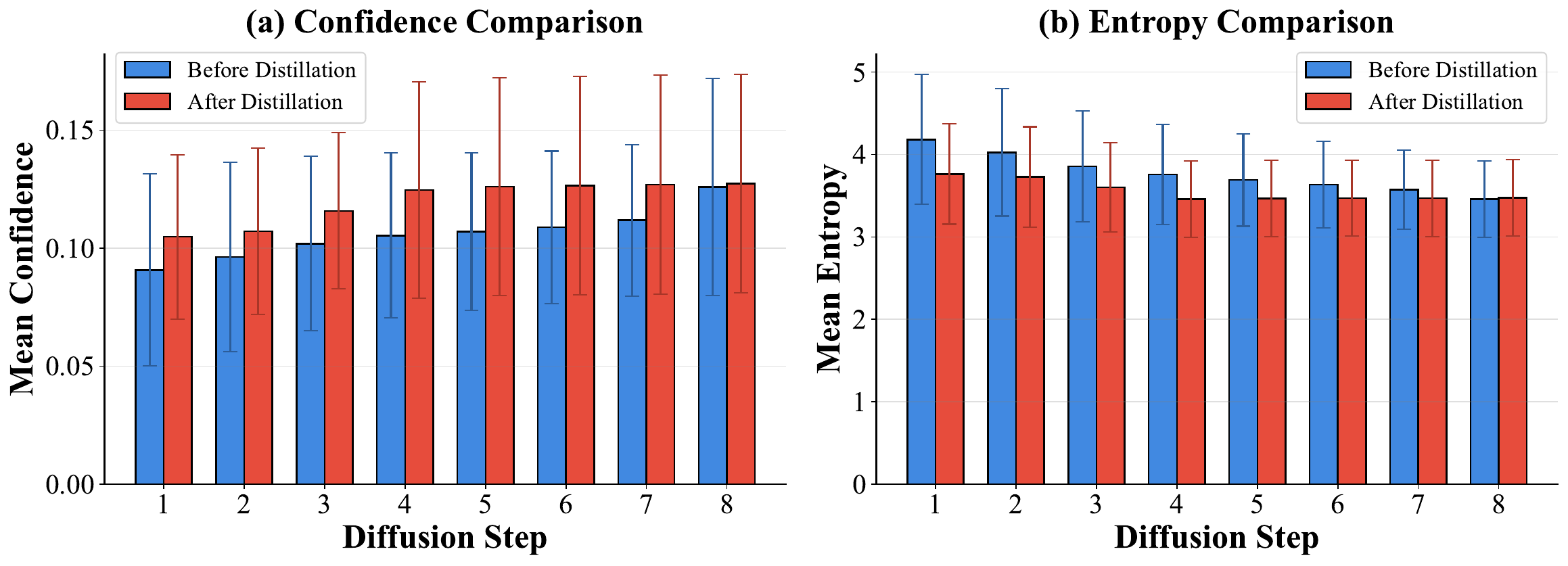}
  \caption{Comparison of prediction uncertainty before and after Iterative Self-Distillation. (a) Mean confidence scores across diffusion steps. (b) Mean entropy values across diffusion steps.}
  \label{fig:uncertainty_comparison}
\end{figure}

\subsection{Impact of Iterative Self-Distillation}
\paragraph{Hyperparameter Analysis.}
We analyze the impact of distillation loss weight $\alpha$ and temperature $\tau$ on performance (Table~\ref{tab:distillation_hyperparams}). Without distillation ($\alpha=0$), WER reaches 6.29 and 41.27 at Step 4 and Step 1, compared to 5.34 and 6.23 under $\alpha=0.7$ and $\tau=2.0$. This indicates the effectiveness of Iterative Self-Distillation in compressing multi-step refinement. However, excessive distillation weight ($\alpha=0.9$) degrades performance, indicating the need to balance teacher guidance with ground-truth supervision. For temperature, $\tau=2.0$ consistently outperforms $\tau=1.0$ and $\tau=3.0$, as extreme values produce teacher distributions that are either overly sharp or overly smooth. Thus, we adopt $\alpha=0.7$ and $\tau=2.0$ as our default configuration.

\paragraph{Uncertainty Reduction Analysis.}
We further examine how distillation accelerates convergence by measuring prediction uncertainty through mean confidence and entropy. As shown in Figure~\ref{fig:uncertainty_comparison}, distillation substantially reduces uncertainty at early diffusion steps: Step 1 and 2 achieve similar confidence and entropy levels comparable to baseline Step 4, while Step 4 reaches performance matching baseline Step 8. These findings demonstrate that Iterative Self-Distillation effectively transfers the uncertainty resolution process from later steps to earlier ones, enabling few-step generation to match the quality of multi-step generation.

%% file: sections/5_Conclusion.tex
\section{Conclusion}
This work systematically explores MDM as a non-autoregressive paradigm for end-to-end spoken dialogue. We introduce two key mechanisms: Hierarchical Block-wise Masking to mitigate training-inference distribution mismatch in block diffusion decoding, and Iterative Self-Distillation to compress multi-step refinement into few-step inference via parallel distillation. Experiments demonstrate $3.7\times$--$10\times$ decoding speedup with 34\% first-chunk latency reduction compared to AR baselines, while maintaining competitive WER and state-of-the-art text quality and speech naturalness. Our results suggest that carefully designed MDM approaches can achieve superior efficiency-quality trade-offs, opening a new paradigm for low-latency speech LLMs.

%% file: sections/6_Appendix.tex
\section{Related Work}
\subsection{End-to-End Spoken Dialogue Systems}

End-to-end spoken dialogue systems adopt two paradigms: Native Multimodal models and Aligned Multimodal models~\citep{ chen2025minmo}.

Native multimodal models integrate understanding and generation into unified frameworks. Baichuan-Omni-1.5~\citep{li2025baichuan} achieves omni-modal processing via a single backbone trained on large-scale multimodal data. GLM-4-Voice~\citep{zeng2024glm} leverages ultra-low bitrate tokenization and interleaved pre-training for emotive generation. Step-Audio 2~\citep{wu2025step} integrates Reinforcement Learning and Retrieval-Augmented Generation, while MiMo-Audio~\citep{coreteam2025mimoaudio} optimizes tokenization for high-density sequences. Kimi-Audio~\citep{ding2025kimi} unifies continuous acoustic and discrete semantic inputs, and SLAM-Omni~\citep{chen2025slam} utilizes single-stage training with grouped semantic prediction.

Conversely, Aligned Multimodal models prioritize modular alignment to preserve LLM reasoning. Qwen2.5-Omni~\citep{xu2025qwen2} and Qwen3-Omni~\citep{Qwen3-Omni} adopt the Thinker-Talker architecture, with the latter further incorporating Mixture-of-Experts (MoE) for efficiency. MiniCPM-o 2.6~\citep{MiniCPM-o-2.6} focuses on efficient on-device processing. For decoding, LLaMA-Omni~\citep{fang2024llama} uses a Non-Autoregressive CTC decoder for low latency, whereas LLaMA-Omni 2~\citep{fang2025llama} employs an Autoregressive streaming decoder for naturalness. To reduce latency, VITA-Audio~\citep{long2025vita} introduces Multiple Cross-modal Token Prediction (MCTP). VocalNet~\citep{wang2025vocalnet1} employs Multi-Token Prediction (MTP), and VocalNet-M2~\citep{wang2025vocalnet} augments MTP with multi-codebook tokenization.

Despite these innovations, the predominance of Autoregressive generation remains a limitation. Its strict serial dependency conflicts with the inherent redundancy of speech, inducing error accumulation and latency bottlenecks that impede the balance between responsiveness and quality.

\subsection{Masked Diffusion Modeling}

Masked Diffusion Modeling represents a Non-Autoregressive paradigm that breaks serial dependencies by predicting masked tokens in parallel using bidirectional context. Recent advancements have demonstrated its potential in Large Language Models: LLaDA~\citep{nie2025large} demonstrated performance comparable to AR baselines, while subsequent scaling studies~\citep{nie2024scaling} established the laws for MDM. Furthermore, LLaDA 1.5~\citep{zhu2025llada} and LLaDA-MoE~\citep{zhu2025lladamoe} advanced alignment optimization and sparse architectures. Moreover, MDM has been successfully extended to multimodal and domain-specific tasks; LLaDA-V~\citep{you2025llada} and MMaDA~\citep{yang2025mmada} achieve unified vision-language modeling, whereas DiffuCoder~\citep{gong2025diffucoder} and Dream 7B~\citep{ye2025dream} demonstrate superior non-causal reasoning capabilities in code generation and planning.

Hybrid modeling paradigms have also attracted widespread attention. SDAR~\citep{cheng2025sdar} proposes a collaborative diffusion-autoregressive framework that maintains inter-block global coherence while enabling intra-block parallel decoding. Similarly, TtT~\citep{liu2025text} integrates autoregressive text generation with non-autoregressive audio diffusion within a unified spoken dialogue system, implementing modality-specific processing mechanisms.

Despite parallel generation capabilities, MDM deployment is hindered by inference inefficiency, as optimal quality typically necessitates extensive many diffusion steps. This cumulative overhead often negates parallelization gains, making the compression of iteration steps without quality compromise a pivotal challenge.

\subsection{Inference Acceleration and Distillation}

To mitigate the latency overhead of diffusion decoding in MDM, researchers have primarily explored two acceleration directions: KV Cache adaptation and Knowledge Distillation.

KV Cache, while standard in Autoregressive models, poses challenges for the bidirectional attention of MDM. Fast-dLLM~\citep{wu2025fast} introduces a block-wise approximate KV Cache with confidence-aware decoding. dKV-Cache~\citep{ma2025dkv} proposes delayed and conditional caching strategies based on token diffusion dynamics. Furthermore, Fast-dLLM v2~\citep{wu2025fastv2} and D2F~\citep{wang2025diffusion} efficiently convert pretrained AR models into block-diffusion frameworks compatible with standard caching mechanisms. However, while these methods enhance global throughput, they fail to address the Time to First Block. The initial block generation still necessitates a full iterative process without historical caches, thereby impeding instantaneous responsiveness in streaming contexts.

Alternatively, Knowledge Distillation (KD) accelerates inference by compressing the multi-step refinement process. SDTT~\citep{deschenaux2024beyond} employs self-distillation to match the teacher's multi-step distribution, reducing inference steps by orders of magnitude. Di4C~\citep{hayakawa2024distillation} distills dimensional correlations via a hybrid student-teacher architecture. Di[M]O~\citep{zhu2025di} achieves one-step generation by distilling masked diffusion models through an on-policy framework. CDLM~\citep{kim2025cdlm} integrates consistency modeling with block-causal masking to reduce sampling steps. Nevertheless, these frameworks entail complex multi-model coordination. Furthermore, the extensive length of speech sequences incurs prohibitive training costs, collectively rendering existing approaches ill-suited.

\section{Mathematical Formulations}
\subsection{Sparse Semantic Anchor Alignment}
\label{sec:appendix_anchor_properties}

We formalize the intra-block sparse semantic anchor alignment introduced in Section~\ref{sec:method_architecture}, which constructs an aligned semantic sequence $\mathbf{h}'_{1:T}$ from the semantic features $\mathbf{h}_{1:N}$.

Using the block partition defined in Section~\ref{sec:method_mdm} with block index sets $\mathcal{I}_k$ in Equation~\eqref{eq:block_partition}, we define the anchor positions and the corresponding assignment rule below.

\paragraph{Anchor position set.}
Within the $k$-th block, anchors are placed at the first $Q$ positions. Formally, the anchor set in block $k$ is defined as
\begin{equation}
\mathcal{A}_k = \{(k-1)B + q \mid q \in \{1, 2, \ldots, Q\}\} \cap \mathcal{I}_k,
\label{eq:anchor_block}
\end{equation}
and the global anchor set is $\mathcal{A} = \bigcup_{k=1}^{K_{\text{blk}}} \mathcal{A}_k$, 
with sorted positions $\{a_1 < a_2 < \cdots \}$.

\paragraph{Order-preserving semantic assignment.}
For each anchor position $a_m$ in the sorted sequence, we assign semantic features sequentially:
\begin{equation}
\mathbf{h}'_{a_m} = \begin{cases}
\mathbf{h}_m, & \text{if } m \leq N, \\
\mathbf{0}, & \text{if } m > N.
\end{cases}
\label{eq:anchor_assignment}
\end{equation}
For non-anchor positions $t \notin \mathcal{A}$, we set $\mathbf{h}'_t = \mathbf{0}$. 

This assignment preserves the sequential correspondence between anchor indices and semantic indices, as $a_m < a_{m'}$ implies $m < m'$.

\subsection{Hierarchical Block-wise Masking Strategy}
\label{sec:appendix_masking}
This section provides the algorithmic details of the Hierarchical Block-wise Masking strategy introduced in Section~\ref{sec:method_mdm}. Algorithm~\ref{alg:hier_mask} presents the complete sampling procedure.

\paragraph{Expected masking ratio.}
Ignoring boundary effects in the final block and approximating all block lengths by $B$, the expected number of masked tokens can be written as
\begin{equation}
\mathbb{E}[|\mathcal{M}|] \approx K_{\text{blk}} \cdot \mathbb{E}[\gamma_c] \cdot B \cdot \mathbb{E}[\gamma_t].
\end{equation}
\paragraph{Comparison to Global Bernoulli masking.}
To characterize intermediate partially-observed states within a block, we define the intra-block masking ratio for block $k$ as
\begin{equation}
R_k \triangleq \frac{|\mathcal{M}\cap \mathcal{I}_k|}{|\mathcal{I}_k|}.
\end{equation}
For any selected block $k\in\mathcal{K}_{\text{mask}}$, Algorithm~\ref{alg:hier_mask} samples $n_k$ masked positions inside the block, hence $R_k=n_k/|\mathcal{I}_k|$. When $|\mathcal{I}_k|=B$ and $\gamma_t\ge 1/B$, we have $n_k=\lfloor \gamma_t B\rfloor$, which implies
\begin{equation}
0 \le \gamma_t - R_k < \frac{1}{B}.
\end{equation}
Since $\gamma_t \sim U(\gamma_t^{\min},\gamma_t^{\max})$, the inequality above shows that, within selected blocks, training covers a family of masking ratios $R_k$ over $[\gamma_t^{\min},\gamma_t^{\max}]$ with resolution $1/B$. This directly matches the range of intermediate states where a block transitions from highly masked to lightly masked during block diffusion decoding.

\begin{algorithm}[t]
\caption{Hierarchical Block-wise Masking Sampler}
\label{alg:hier_mask}
\begin{algorithmic}[1]
\Require Sequence length $T$, block size $B$, ratio ranges $[\gamma_c^{\min},\gamma_c^{\max}]$ and $[\gamma_t^{\min},\gamma_t^{\max}]$
\Ensure Masked index set $\mathcal{M}\subseteq\{1,\dots,T\}$
\State $K_{\text{blk}} \gets \lceil T/B\rceil$; $\mathcal{M}\gets \emptyset$
\State Sample $\gamma_c \sim U(\gamma_c^{\min},\gamma_c^{\max})$ and set $M_{\text{blk}} \gets \lfloor \gamma_c \cdot K_{\text{blk}}\rfloor$
\State Sample $\mathcal{K}_{\text{mask}} \subseteq \{1,\dots,K_{\text{blk}}\}$ uniformly without replacement s.t.\ $|\mathcal{K}_{\text{mask}}|=M_{\text{blk}}$
\State Sample $\gamma_t \sim U(\gamma_t^{\min},\gamma_t^{\max})$
\For{$k \in \mathcal{K}_{\text{mask}}$}
    \State $\mathcal{I}_k \gets \{t \mid (k-1)B < t \le \min(kB,\,T)\}$
    \State $n_k \gets \max\!\big(1,\lfloor \gamma_t \cdot |\mathcal{I}_k|\rfloor\big)$
    \State Sample $\mathcal{M}_k \subseteq \mathcal{I}_k$ uniformly without replacement s.t.\ $|\mathcal{M}_k|=n_k$
    \State $\mathcal{M} \gets \mathcal{M} \cup \mathcal{M}_k$
\EndFor
\State \Return $\mathcal{M}$
\end{algorithmic}
\end{algorithm}

For comparison, Global Bernoulli masking masks each token independently with probability $\gamma_g$.
For any equal-length block and any $\delta>0$, Hoeffding's inequality and the union bound yield
\begin{equation}
\mathbb{P}\Big(\max_{1\le k\le K_{\text{blk}}}|R_k-\gamma_g|\ge \delta \,\Big|\, \gamma_g\Big)
\le 2K_{\text{blk}}e^{-2B\delta^2}.
\end{equation}
This bound indicates that, as $B$ increases, all blocks in a single sample concentrate around the same global ratio $p$ with high probability. While this property provides uniform and comprehensive training signals suitable for foundational MDM learning, our Hierarchical Block-wise sampler explicitly induces diverse intra-block ratios $R_k$ in selected blocks, which more directly simulates the intermediate states encountered during block diffusion decoding.

\section{Forward and Reverse KL for Iterative Self-Distillation}
\label{sec:appendix_kl_analysis}

Given two distributions $P_{\text{tea}}$ and $P_{\text{stu}}$ defined over vocabulary $\mathcal{V}$, the asymmetry of KL divergence leads to fundamentally different behaviors when optimizing student parameters. The gradients of forward KL (mean-seeking) and reverse KL (mode-seeking)~\citep{gu2023minillm,wu2025rethinking} are respectively:
\begin{equation}
\begin{aligned}
\nabla_\theta \mathrm{KL}(P_{\text{tea}}\|P_{\text{stu}}) &= -\sum_{v\in\mathcal{V}} P_{\text{tea}}(v) \\
&\quad \nabla_\theta \log P_{\text{stu}}(v;\theta),
\end{aligned}
\end{equation}
\begin{equation}
\begin{aligned}
\nabla_\theta \mathrm{KL}(P_{\text{stu}}\|P_{\text{tea}}) &= \sum_{v\in\mathcal{V}} P_{\text{stu}}(v;\theta) \\
&\quad \nabla_\theta \log\frac{P_{\text{stu}}(v;\theta)}{P_{\text{tea}}(v)}.
\end{aligned}
\end{equation}
Forward KL is weighted by $P_{\text{tea}}(v)$, requiring the student to cover all non-zero regions of the teacher; reverse KL is weighted by $P_{\text{stu}}(v)$, allowing the student to focus on high-probability regions.

In Iterative Self-Distillation, later-step teacher predictions are typically more informative but can remain multi-modal, especially for ambiguous tokens. In this setting, reverse KL is often preferable because it encourages the student to concentrate on the teacher's high-probability regions rather than allocating extra mass to low-probability tails, whereas forward KL tends to be more mean-seeking and can over-emphasize covering all modes~\citep{gu2023minillm,wu2025rethinking}.

In our task, the teacher distribution after $K$ iterations is more refined yet retains uncertainty. Forward KL forces the student to spread mass across sub-optimal candidates where the teacher remains uncertain, while reverse KL allows the student to concentrate on regions the teacher progressively determines. From the gradient perspective, reverse KL imposes smaller penalties on low-probability regions weighted by $P_{\text{stu}}(v)$, whereas forward KL significantly penalizes even in these regions, causing resource waste. This aligns with the objective of simulating multi-step refinement in a single pass.

Table~\ref{tab:kl_comparison} compares the distillation performance of both KL variants. Results show that reverse KL achieves superior performance under low-step inference, validating its compatibility with progressive refinement tasks.

\begin{table}[t]
\centering
\resizebox{\columnwidth}{!}{%
\begin{tabular}{lccc}
\toprule
\textbf{Model} & \textbf{KL Type} & \textbf{WER (\%)} $\downarrow$ & \textbf{UTMOS} $\uparrow$ \\
\midrule
\multirow{2}{*}{VocalNet-MDM (Step 4)} 
& Forward KL  & 6.41 & 4.48 \\
\cdashline{2-4}
& Reverse KL  & \textbf{5.34} & \textbf{4.49} \\
\midrule
\multirow{2}{*}{VocalNet-MDM (Step 2)} 
& Forward KL  & 7.41 & 4.46 \\
\cdashline{2-4}
& Reverse KL  & \textbf{6.10} & \textbf{4.47} \\
\midrule
\multirow{2}{*}{VocalNet-MDM (Step 1)} 
& Forward KL  & 12.14 & 4.40 \\
\cdashline{2-4}
& Reverse KL  & \textbf{6.23} & \textbf{4.46} \\
\bottomrule
\end{tabular}%
}
\caption{Comparison of forward KL and reverse KL in Iterative Self-Distillation.}
\label{tab:kl_comparison}
\end{table}

\section{Evaluation Details}
\label{sec:appendix_eval}
\begin{figure}[h]
  \centering
  \includegraphics[width=0.48\textwidth]{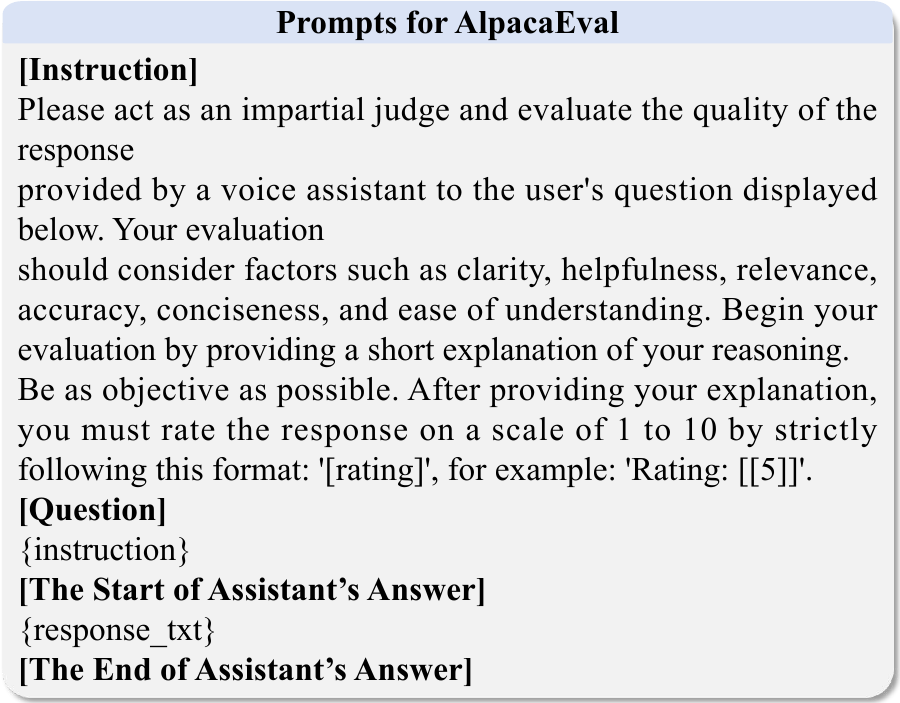}
  \caption{Prompt for AlpacaEval.}
  \label{fig:prompts_1}
\end{figure}

\begin{figure}[h]
  \centering
  \includegraphics[width=0.48\textwidth]{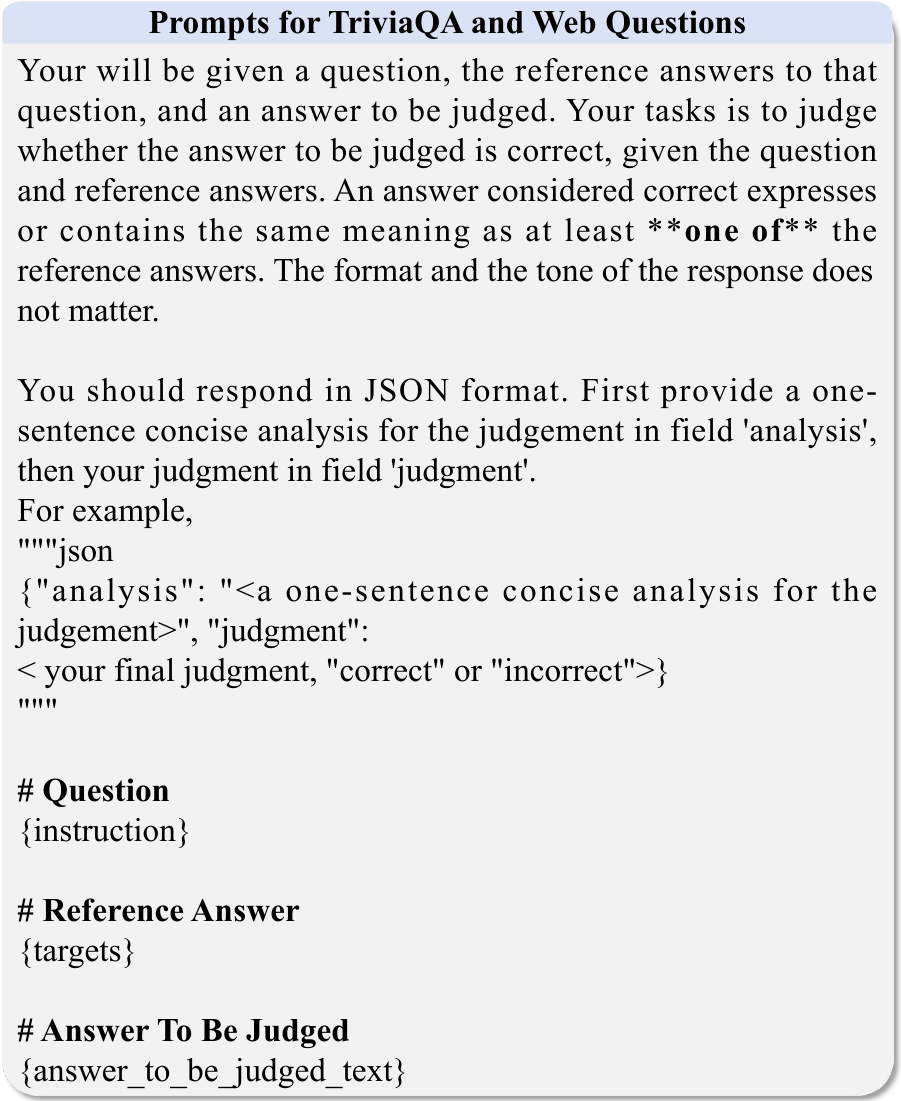}
  \caption{Prompt for TriviaQA and Web Questions.}
  \label{fig:prompts_2}
\end{figure}

\begin{figure}[h]
  \centering
  \includegraphics[width=0.48\textwidth]{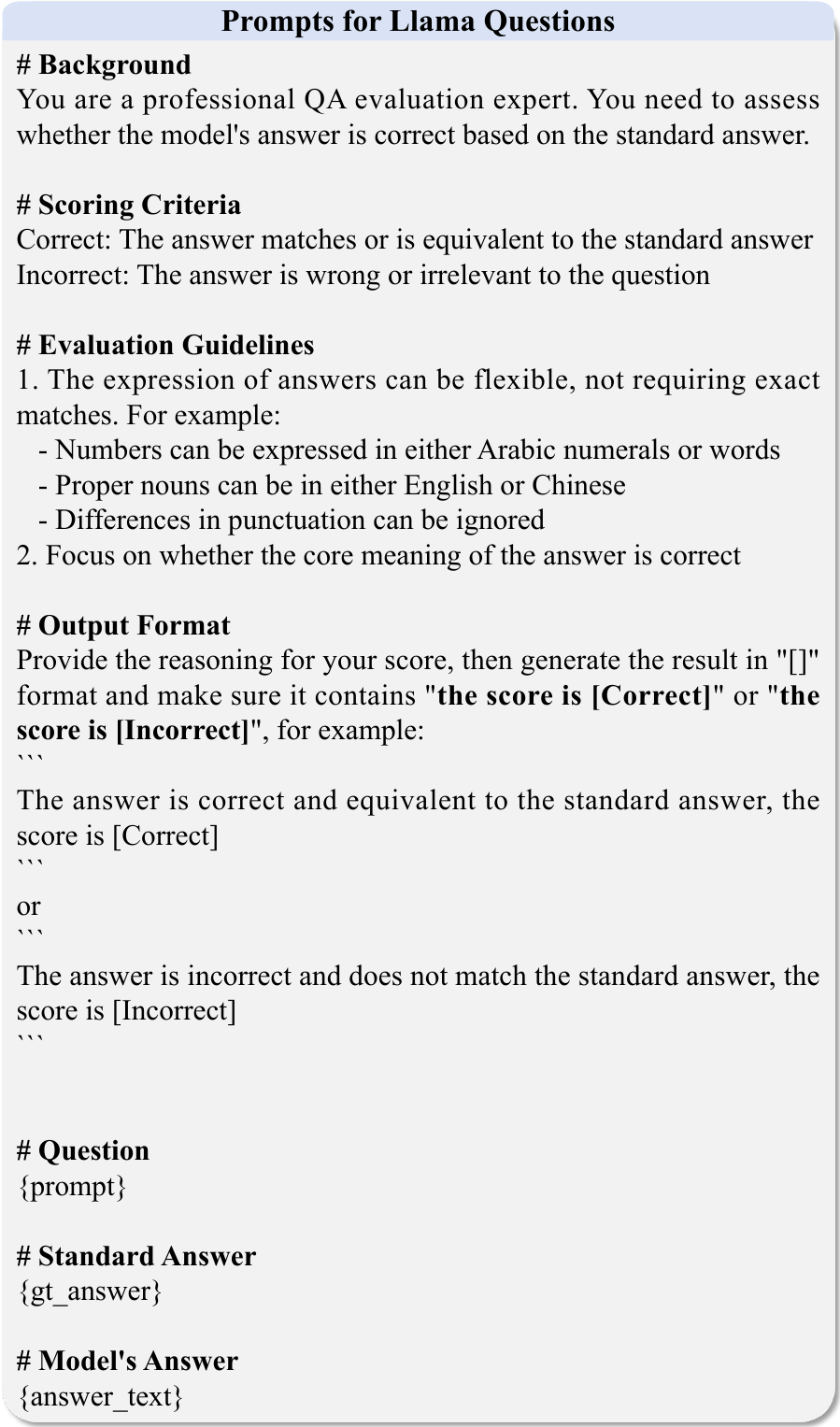}
  \caption{Prompt for Llama Questions.}
  \label{fig:prompts_3}
\end{figure}

To assess model performance, we adopt an LLM-as-a-Judge evaluation paradigm and use the evaluation prompt templates provided by OpenAudioBench \cite{li2025baichuan} to score model outputs in a unified manner. Our evaluation covers two task settings: open-ended generation and semi-open question answering. Specifically, the open-ended setting uses AlpacaEval, while the semi-open QA setting includes Llama Questions, TriviaQA, and Web Questions. For semi-open QA tasks, we include reference answers in the evaluation prompts to provide explicit judging criteria and to assist the judge LLM in determining the correctness of model responses. For open-ended tasks, the judge LLM assigns holistic scores according to multiple qualitative aspects, including clarity, helpfulness, relevance, accuracy, conciseness, and ease of understanding. The complete prompt designs and scoring guidelines are provided in Figures~\ref{fig:prompts_1}, \ref{fig:prompts_2}, and \ref{fig:prompts_3}.

